\documentclass{article}

\PassOptionsToPackage{numbers,sort&compress}{natbib}
\usepackage[final,main]{neurips_2026}

\usepackage[utf8]{inputenc}
\usepackage[T1]{fontenc}
\usepackage{url}
\usepackage{booktabs}
\usepackage{amsfonts}
\usepackage{microtype}
\usepackage{xcolor}
\usepackage{graphicx}
\usepackage{array}
\usepackage{float}
\usepackage{longtable}
\usepackage{tabularx}
\usepackage{etoolbox}
\usepackage{hyperref}

\newcolumntype{Y}{>{\raggedright\arraybackslash}X}

\makeatletter

\renewcommand{\@notice}{}
\patchcmd{\LT@output}{\vss}{\vfil}{}{\PackageError{paper}{Longtable pagination patch failed}{Check the installed longtable version.}}
\patchcmd{\LT@output}{\vss}{\vfil}{}{\PackageError{paper}{Longtable pagination patch failed}{Check the installed longtable version.}}
\makeatother

\title{Silent Failures in Agent--Tool Interaction: An Audit of ToolUniverse}

\author{%
  Devansh Singh \\
  AI Tech Ethics \\
  \texttt{devansh.singh@aitechethics.com}
  \And
  Shreya Gopalan \\
  AI Tech Ethics \\
  \texttt{shreya.gopalan@aitechethics.com}
  \And
  Sundaraparipurnan Narayanan \\
  AI Tech Ethics \\
  \texttt{sundar.narayanan@aitechethics.com}
}

\begin{document}
\raggedbottom

\maketitle

\begingroup
\renewcommand{\thefootnote}{}
\makeatletter
\let\@footnotetext\H@@footnotetext
\footnotetext{{\fontsize{10}{12}\selectfont
Accepted at the GenAI4Health and AgenticLS workshops at NeurIPS 2026.}}
\makeatother
\endgroup

\begin{abstract}
Agentic AI systems are increasingly adopting automated pipelines that integrate multiple tools. While prior research and benchmarks have studied about task success and task completion of these agentic systems, research on agent to tool interaction chain via a wrapper and API, specifically in biological agentic workflows, remains limited. This study investigates specific failures in agent to tool interaction where a tool invocation appears successful but some or all of the information or functionality from the tool via its API / wrapper is incomplete or missing and there are no communications / notifications to the user or the agent regarding the missing information. We call these 'silent failures' because the user or the agents are not aware that such a failure has occurred. For the purpose of this study, we developed an audit mechanism to identify such silent failures in agent to tool interaction chains, by examining 15 scientific tools (and their associated API documentation and tool documentations) integrated within the ToolUniverse environment (ToolUniverse serves as our experimental environment rather than the object of the study itself). We structure our study around 7 failure locus, characterising where the failure occurs in the chain. We observed 91 failures (manually validated post LLM based candidate discovery and automated testing), most frequent of them being missing data or fields and inconsistencies in search, filtering or ranking criteria. Most of the 91 failures occurred in API layer (52) or wrapper layer (25), with a potential of silent failure amplification downstream. The results show that silent failures originate upstream of the event and propagate downstream into apparently valid scientific outputs. We propose a concept of contextual reliability to handle such failures and suggest mechanisms for testing, disclosing, monitoring, and measuring such failures across the agent--tool interaction pipeline. 
\end{abstract}

\section{Introduction}
\paragraph{Agent-Tool interaction:} AI systems for science are increasingly connecting large language models (LLMs) and agents to domain-specific databases, wrappers or APIs, computational or other specialized tools, and ever growing literature, to provide stronger research and solutions for domain-specialists. Such systems in biology including Virtual Lab, ToolUniverse, and Biomni demonstrate the use of tools within agentic scientific workflows \citep{swanson2025virtual,gao2025tooluniverse,huang2025biomni,gxl2026paperclip}. In such systems, the agents integrated with large multimodal models, not only handle multi-modality of data, but also search, select, collaborate, collate, interpret, and combines external scientific information and tools. However, the reliability of the interaction between an agent and these tools remains insufficiently evaluated. Existing evaluations of agentic systems commonly emphasize task outcome success including resolving an issue in SWE-bench, completing an interactive task in AgentBench, or executing a valid tool-use trajectory in ToolBench \citep{jimenez2024swebench,liu2024agentbench,qin2024toolllm}. These evaluations do not necessarily establish whether the information returned by the tool is complete, appropriately qualified, correctly represented, or correctly interpreted by the agent. A tool may return a valid response while the final answer is incomplete, incorrectly qualified, or scientifically misinterpreted. This problem is particularly consequential in biology, where provenance, evidence strength, experimental status, organism and disease context, and completeness can determine the meaning of a result. At scale, interactions that cannot be individually inspected accumulate verification debt \citep{mo2026age,taktakidze2026paradigm,vukojevic2026cheap}.We choose, information flow from the agent to tool interaction pipeline. In specific, we were interested in understanding where the pipeline fails and the message stays silent for the agent or for the user. Prior work on LLM-agent debugging shows that errors can cascade across planning, memory, action, and system-level components \citep{zhu2025fail}; provenance related work also highlight tracing data and transformations of such data across the scientific workflows \citep{buneman2001why,simmhan2005survey,davidson2008provenance}. We refer to the subset of such errors that remain undisclosed as \emph{silent failures}. 

\paragraph{Defining silent failure in agent-tool interaction:} We define a silent failure as an interaction in which (1) the tool invocation appears successful, (2) the returned information is incomplete, transformed, ambiguous, or otherwise insufficient for the intended task, and (3) the limitation is not adequately disclosed to the downstream user or agent. Unlike a loud failure, where an API timeout reported to the user, a silent failure can arise when an API returns its first 1,000 records (out of 10000 records) and the agent reports ``the records.'' Examples include a truncated result presented as complete results, an unavailable filter information silently generalized by an agent, a missing qualifier treated as absence of information, and a score or evidence hierarchy interpreted incorrectly. Silent failures can therefore propagate across system layers. An upstream limitation that is relatively benign at the tool or API layer can become a more consequential error when the agent interprets the partial output as complete evidence. We refer to this as silent amplification.

\paragraph{Research theme: } Our focus is distinct from general studies of an agent's internal reasoning failures or its final task success. We audit the \emph{agent--tool interaction}: whether the agent, wrapper, API, and underlying resource preserve the information, qualifiers, provenance, scope, and semantics needed for the agent's claim. Thus, even an agent that plans correctly and completes a task can silently amplify a limitation introduced at an external tool boundary. These risks are especially consequential in biology, where downstream reuse and interpretation depend on rich, contextual metadata and provenance \citep{wilkinson2016fair}.

\begin{figure}[H]
 \centering
 \fbox{\parbox{0.95\linewidth}{\centering
  User intent $\rightarrow$ Agent $\rightarrow$ Wrapper $\rightarrow$ API $\rightarrow$ Tool/Data $\rightarrow$ API $\rightarrow$ Wrapper $\rightarrow$ Agent $\rightarrow$ Claim}}
  \caption{The agent--tool interaction chain. A response can be technically successful while information, qualifiers, provenance, scope, or meaning is lost at one or more boundaries.}
  \label{fig:interaction-chain}
\end{figure}

Our central question is: \emph{when an agent successfully uses a scientific tool, how reliably does the interaction preserve the information, qualifiers, context, and meaning required to support the intended conclusion?}. 

Our research questions are: 
\begin{itemize}
  \item \textbf{RQ1:} How prevalent are potential silent failures when scientific tools are used through an agentic interface?
  \item \textbf{RQ2:} At which locus in the Agent--Wrapper--API--Tool chain do these failures occur?
  \item \textbf{RQ3:} Which information, qualifier, context, and semantic properties are most vulnerable?
  \item \textbf{RQ4:} Does the agent preserve, disclose, or amplify upstream limitations?
\end{itemize}

We address these research questions by examining tools integrated within ToolUniverse environment.

\paragraph{Failure loci in agent-tool interactions: } We organize our research in Failure Locus (the point at which the failure originates thematically). Our failure locus taxonomy consists of the following: (L1) \emph{Tool Limitations:} inherent limitation of the tool or data or meaning associated with the tool; (L2) \emph{API Gap:} tool/webUI has the capability/ the feature, however, it is completely missing from the API (no endpoint / no way to call it); (L3) \emph{API Partial Gap:} tool/webUI has the capability/ the feature, however, it is partially missing from the API. For instance, the endpoint exists, but certain fields are not covered as part of the API in comparison to the tool; (L4) \emph{Wrapper Gap:} tool/webUI and the API has the capability/ the feature, however, no wrapper exposes such capability; (L5) \emph{Wrapper Partial Gap:} tool/webUI and the API has the capability/ the feature, however, wrapper partially omits parameters, fields, or formats which the API returns; (L6) \emph{Agent Usability Gap:} Wrapper provided a usable output, but the agent could not use it; and (L7) \emph{Agent Interpretation Gap:} The AI agent received the wrapper output but misinterpreted it. This includes overstating, understating, treating incomplete as complete or vice versa.

\section{Experimental environment and methodology}

\begin{figure}[H]
  \centering
  \makebox[\linewidth][c]{\includegraphics[width=1.1\linewidth]{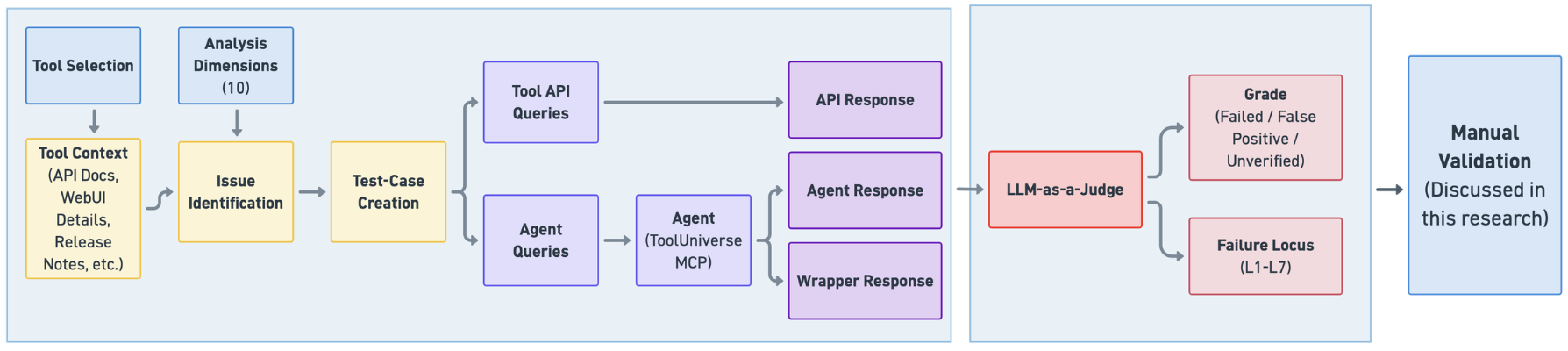}}
  \caption{Flowchart of experimental methodology.}
  \label{fig:flowchart}
\end{figure}

\paragraph{ToolUniverse setup and selection of tools:} ToolUniverse was set up using the official MCP guidance with the AntiGravity CLI agent (using Gemini 3.5 flash lite) as the experimental environment. We randomly sampled 63 main tools having over 285 associated tools in ToolUniverse. In this study, \emph{main tool} refers to the underlying biological database or computational resource and the individual ToolUniverse wrapper or function is called as \emph{tool} or \emph{wrapper}. From the initial sample, 15 tools (having 104 associated ToolUniverse tools) were selected for detailed analysis based on the identification of documented candidate failures that could be subjected to detailed investigation.

\paragraph{Failure candidate discovery:} For each sampled tool, we assembled a tool context containing available documentation and implementation information. The context was provided to the Cursor agent (using Gemini 3 Flash) as a combination of URLs, extracted content, and structured JSON. Sources included tool documentation; web UI details; API documentation; user manual: manual, tool guide, beginner's manual, etc; release notes for both the API and web UI (if available); reported GitHub issues. 

Based on this context, the Cursor agent generated potential failures by identifying conflicts, contract disagreements, implementation issues, and Web/API/Wrapper inconsistencies. The systematic comparison considered functionality, fields, parameters, filtering, ranking, pagination, formats, semantics, provenance, and temporal changes. To address our central question we examined properties like - requested fields / results are present / available; conditions attached to the information are preserved; source/evidence attribution is retained; boundaries of the retrieved result are clear; and the semantic interpretation of the returned information / outcome is preserved. Preserving meaning is as important as preserving the information. For instance, collapsing phospho-SMAD3 to SMAD3, flattening CIViC evidence and assertions, or exposing a low-level genome-quality flag without its specific interpretation can impact the conclusions arrived by the agent. 

We also examined \textbf{evolutionary context}, including changes in APIs, Web UIs, documentation, and tool releases that could alter data scope, assumptions, interpretation, or meaning. This was used to identify candidate failures involving temporal changes or changes in the context in which returned information should be interpreted.

\paragraph{Test-case and query generation: } For each candidate failure, concrete test cases and queries were generated to determine whether the observed limitation could be reproduced through ToolUniverse. Test-case and query generation was iteratively refined through manual verification before automating. 
\begin{itemize}
\item \textbf{Tool-level specification:} Queries identified the underlying main tool rather than prescribing a specific ToolUniverse tool, allowing the agent to select the relevant functionality.
\item \textbf{Exact tool names:} Tool names were restricted to names present in ToolUniverse.
\item \textbf{Complete inputs:} Each test case contained the specific identifiers, parameters, or other inputs required to execute the test. For example, a test case specified an accession such as \texttt{DP00086} rather than an underspecified entity such as TP53.

\item \textbf{Concrete scientific inputs:} Test cases used concrete biological entities, identifiers, datasets, or other domain-specific inputs rather than hypothetical examples.

\item \textbf{Failure isolation:} Queries were designed to test the target failure rather than ask the agent to independently determine whether a failure existed. In particular, queries did not instruct the agent to compare its output against the raw API, assess completeness, or simulate infrastructure failures to separate confounding between the target tool failure and failure introduced by the test query
\end{itemize}

For example, a test query could request: "Use the ToolUniverse tool \texttt{DisProt\_get\_entry} with \texttt{accession='DP00086'}. What is the disorder score of each residue and how can we interpret the score in terms of the whole protein? Also list every returned field in the tool output."; rather than asking the agent to determine whether the output was complete or to compare it against an external source. We choose this distinction because the objective of the test was to observe the behavior of the Agent--Tool interaction, rather than to delegate failure diagnosis to the agent itself.

\paragraph{Repeated experimental runs:} Each candidate failure was evaluated through repeated agent executions. An initial test query was generated for each candidate. If the initial execution was classified as a failure, four additional queries targeting the same failure were executed. If the initial execution was classified as false positive or unverified, two additional queries were executed and judged. Additional queries used different concrete scientific inputs where possible to assess whether the observed behavior generalized beyond a single test case. 

\paragraph{Failure validation and locus attribution:} The Cursor agent was used as an automated evaluator of the execution traces. For each run, it classified the result as either \emph{Fail} (the candidate failure was observed) or \emph{False Positive} (the candidate failure was not observed). When the evaluator could not establish either outcome from the available evidence, the case was marked \emph{Unverified / Needs Additional Validation}. Each validated failure was subsequently mapped to the failure-locus taxonomy defined earlier in section 1, distinguishing tool limitations, API gaps, wrapper gaps, agent usability failures, and agent interpretation failures. 

\paragraph{Manual validation:} While the automated audit was performed for 63 \emph{main tools}, two authors manually reviewed 15 randomly-selected main tools (198 test cases) to validate the LLM-driven process. The author tested reviewed the outcomes from cursor evaluation and reperformed them using API and ToolUniverse-Antigravity-CLI agent calls to validate the automated evaluations. The results represented in the report covers only the human verified outcomes from the 15 main tools. Manual validation of the failure test cases involved checking the failure test cases for the following: (1) \emph{Fact Checking} : Is the selected candidate appropriate for the given failure scenario? This was assessed by manually validating the candidate through analyzing the availability of results for the candidate in the Web UI. (2) \emph{Judgement Validation} : Is the LLM judgement valid for the given failure scenario, and are there other possibilities that the LLM might not have understood in terms of the actual tool functionalities and features, which might contribute to the current judgement rather than an actual failure? 

\paragraph{Failure measurement and rubrics: } Each failure was classified along the following 'analysis dimensions': completeness, correctness, appropriateness, consistency, reliability, attribution, interpretation, relevance, context, and purpose alignment. Appendix~\ref{app:rubrics} describes the dimensions and rubrics in detail.

\section{Results}

Failures were distributed across multiple layers of the Agent--Wrapper--API--Tool interaction. The following analysis summarizes the distribution of confirmed failure cases across the seven failure loci, normalized within each tool. The loci correspond to the failure taxonomy defined in Section 1.

\begin{figure}[H]
  \centering
  \makebox[\linewidth][c]{\includegraphics[width=1.2\linewidth]{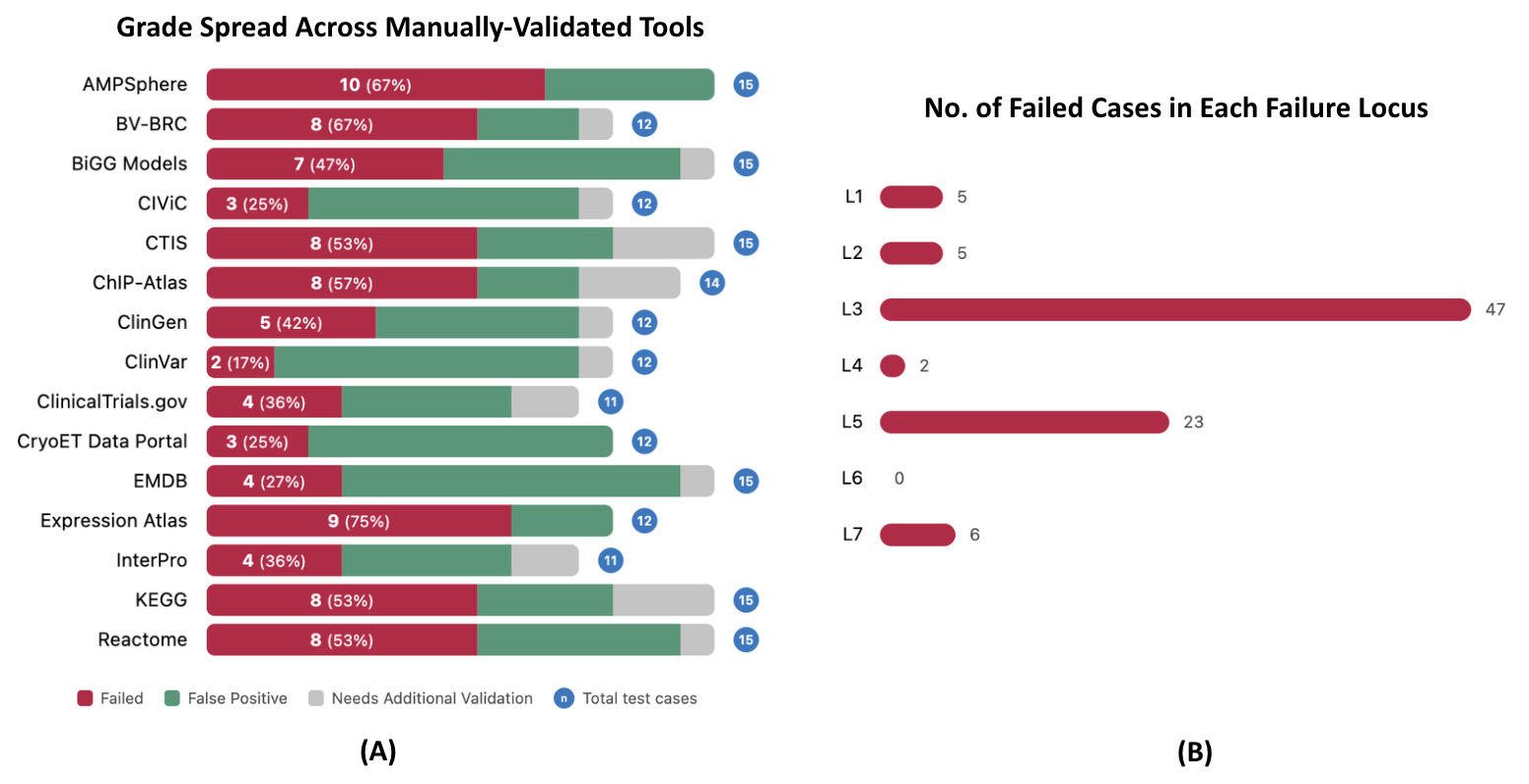}}
  \caption{(A) Distribution of grades across each manually-validated tool. 'Failed' cases are confirmed failures while the issue was not reproduced in 'False Positive' cases. (B) Distribution of failures across each failure locus (failure locus of 3 cases could not be determined).}
  \label{fig:failure-locus-distribution}
\end{figure}

As shown in Figure~\ref{fig:failure-locus-distribution}, 7 out of 15 tools exhibit confirmed failures on more than 50\% of test cases - ExpressionAtlas (75\%), BV-BRC (67\%),  AMPSphere (67\%), CTIS (53\%), ChIP-Atlas (57\%), KEGG (53\%) and Reactome (53\%). The dominant failure loci for most tools was API partial gap (L3) and wrapper partial gap (L5). The dominant failure loci for most tools were API partial gap (L3) and wrapper partial gap (L5). These two loci accounted for the largest \emph{share of human-verified failures} in BiGG Models (100\%), BV-BRC (100\%), CIViC (100\%) and Reactome (88\%). These results indicate that many failures did not arise because an API was entirely unavailable, but because the API provided only a partial representation of the underlying resource, including limitations in fields, filtering, pagination, coverage, or other capabilities. This indicates that partial functionality can produce a valid response while restricting the information available to the agent. The interaction can therefore appear successful even though the returned information is insufficient for the intended scientific conclusion. When the agent treats this response as sufficient evidence, the upstream limitation can be converted into a downstream scientific claim. We refer to this propagation as silent amplification. Other tools also had confirmed failure locus outside L3 and L5. Tool limitations (L1) accounted for 25\% of confirmed cases in ChIP-Atlas, 20\% in AMPSphere. API gap (L2) accounted for 25\% in both CTIS and KEGG. Wrapper complete gap (L4) appeared in KEGG (13\%) and AMPSphere (10\%). Agent interpretation gap (L7) appeared in CryoET Data Portal (33\%), ClinicalTrials.gov (25\%), ChIP-Atlas (13\%).

Failures at the agent-level loci were comparatively uncommon in this analysis. Agent Usability Gap accounted for no confirmed failures in any of the 15 tools, while Agent Interpretation Gap (L7) appeared only in AMPSphere (10\%), ChIP-Atlas (13\%), ClinicalTrials.gov (25\%), CryoET Data Portal (33\%), CTIS (13\%), and KEGG (13\%). This distribution should not be interpreted as evidence that agent-level failures are absent. Rather, within the human vertified cases identified in this study, most failures could be traced to limitations or gaps upstream of agent interpretation. Also the failures are deeply tied to the test cases generated. The volume of generated test cases are based on the failure candidate discovery and hence, the proportion of failures and absence thereof need to be considered in context of such failure candidate discovery process. We emphasise that the results are not representative of exhaustive failures in agent-tool interaction within the ToolUniverse environment for the identified 15 tools, but a illustrative representation of silent failures that exists in such environment. 

\begin{figure}[H]
  \centering
  \includegraphics[width=\linewidth]{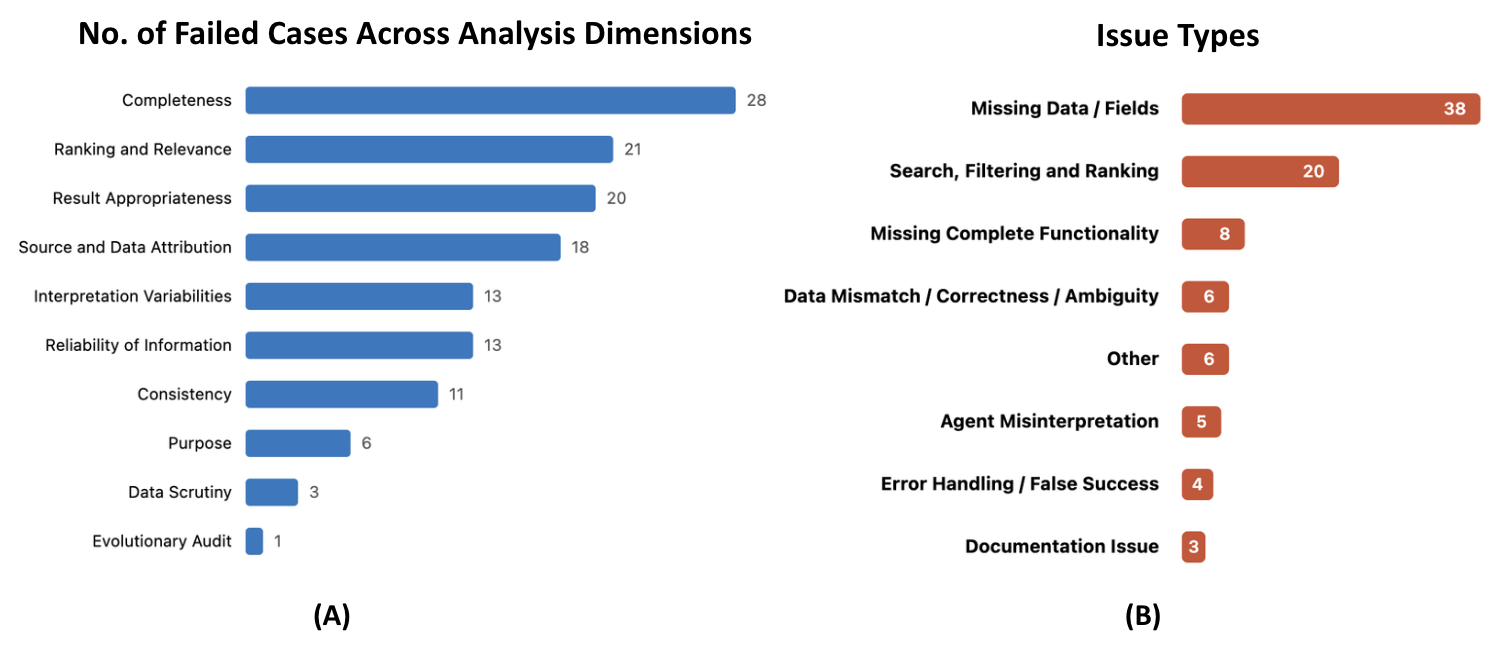}
  \caption{(A) Issue occurrences across the analysis dimensions. Individual test cases may contribute to more than one dimension. (B) Observed types of issues. Individual test cases may exhibit more than one issue.}
  \label{fig:dimension_spread}
\end{figure}

Figure~\ref{fig:dimension_spread} shows the distribution of issue occurrences across the analysis dimensions. The 91 evaluated cases generated 134 issue occurrences because individual test cases could involve multiple dimensions. Completeness was the most frequently affected dimension, with 28 occurrences, followed by Ranking and Relevance (21) and Result Appropriateness (20). Source and Data Attribution (18), Interpretation Variabilities (13), Reliability of Information (13), and Consistency (11) were also frequently represented. The concentration of issues in completeness indicates that failures frequently involve limitations in the amount, scope, or coverage of information available to the agent rather than complete failure to retrieve a result. Interpretation and result appropriateness were the next most frequent dimensions, indicating that obtaining information does not necessarily ensure that it can be appropriately interpreted or used for the intended scientific task.

In terms of types of issues (Fig 4B), missing data/fields were most frequent issue types with 38 occurences, indicating that failures often involved incomplete information rather than complete retrieval failure. Also, search, filtering, and ranking criteria were the second most frequent occurences (20 instances) exhibiting the issues contributed by qualifiers and limitations in retrieval of the complete information. Missing complete functionality (8) and data mismatch/correctness/ambiguity (6) were also observed, showing that available interfaces did not always provide the functionality or representations required by the test case. Agent misinterpretation (5) and error handling/ false success declarations (4) were less frequent but were relevant as they modulate silent failures into confident result outcomes. 

The most consequential cases are propagating phenomena: a limitation at one boundary changes the claim at another. EMDB illustrates API capping becoming a completeness claim; ClinicalTrials.gov illustrates missing filtering becoming inappropriate inclusion; ChIP-Atlas illustrates score or biological-state representation becoming semantic misinterpretation; and CIViC illustrates evidence-hierarchy flattening becoming epistemic confusion. In each case, the agent can amplify rather than merely inherit the limitation. As an additional qualitative observation, we note that the agent generally discloses loud failures (like authorization failures / API unavailability issues) in a single-tool test but it often fails to disclose silent failures (like possible truncation of results due to wrapper limits).

\section{Representative examples}

\paragraph{L1: Tool Limitation} The AMPSphere tool contains a large number of "Blanks" and "Others" without further breakdown into their classification. This ambiguity in data attribution might cause interpretation errors in the agentic workflows. In the same tool, there is a tool-level misclassification of "Progenome" as a habitat, while it is actually a genomic reference database. The search mechanism in the ChIP Atlas tool does not return exact matches, instead it returns similar matches. In one of our test cases, we found that this particular search mechanism cannot differentiate between a molecule and its Post-translation version, leading to irrelevant results returned to the agent. 

\paragraph{L2: API Gap} The ChIP Atlas tool has a Differential Peak analysis tool in the web interface, but there is no API endpoint for that functionality, leading to a case of complete loss of functionality. Similarly, the official KEGG REST API (documented at rest.kegg.jp) provides fundamental lookup, list, and conversion operations but completely lacks endpoints for the high-level 'KEGG Mapper' suite (Reconstruct, Color, Join) mentioned in the WebUI/Tool documentation. The failure manifests at the API layer as a total absence of this functionality, preventing the ToolUniverse wrappers from exposing it and the agent from executing it.

\paragraph{L3: API Partial Gap} The official BiGG Models API endpoint for individual model details returns a null value for the 'organism' field for the model 'iRC1080', despite the organism information being available and displayed in other parts of the BiGG web interface (such as search results). This indicates the API provides incomplete data contract coverage for model details. The AMPSphere tool contains information on the existence of antimicrobial and hemolytic activity probabilities for antimicrobial peptides in the Web UI. However, the official tool API omits these fields entirely from the response, meaning that an agent accessing this tool for an AMP would have no idea of the predicted bioactivity of these peptides.

\paragraph{L4: Wrapper Gap} The functional annotation capability (BlastKOALA/GhostKOALA) is explicitly advertised as a core analysis tool in the KEGG documentation, but it is absent from the KEGG REST API (rest.kegg.jp) documentation and endpoints, preventing the agent from performing sequence-based KO assignment programmatically. In another instance in the AMPSphere tool, there is no dedicated tool for global statistics like family counts and metagenome totals, while the raw API baseline (statistics) provides these values directly. This confirms a L4 gap where core database scale metadata is missing from the ToolUniverse interface. 

\paragraph{L5: Wrapper Partial Gap} The raw API response of the BVBRC tool contains many external and internal IDs (assembly\_accession, bioproject\_accession, p2\_genome\_id), but the wrapper filters these out, returning only the internal genome\_id and taxon\_id. This forces the agent to rely on internal BV-BRC identifiers and prevents easy cross-referencing with NCBI. Another instance of missing partial information through the wrapper was observed in the ClinVar tools. The API  response provides the 'ContributesToAggregateClassification' attribute required to identify which submissions contribute to the aggregate. However, the ToolUniverse wrapper 'ClinVar\_get\_submitted\_records' parses the XML and fails to map or expose this field in its output schema, preventing the AI agent from distinguishing contributing vs. non-contributing records. The CIViC API documentation and GraphQL schema (as per the tool\_api\_response) explicitly include MyVariantInfo fields, indicating the capability exists at the API level. However, the wrapper's implementation of the 'civic\_get\_variant' tool omits these fields in its GQL query, confirming an API-layer integration gap where the wrapper's schema contract fails to expose available API capabilities.

\paragraph{L6: Agent Usability Gap} Issues where the agent is not able to use the data returned by a tool have been reported in forums like GitHub. For example, context-window overflow due to large wrapper outputs, and unusable data formats (like image and audio data for text-only, unimodal agents). While we did not find any instances of L6 issues in manual review, we observed individual wrapper responses which were nearly as large as the agent's context window and tools which returned image data in text-format or a url for the image requiring additional code / processing at the agent harness level. We note that multi-tool, multi-step test-cases may expose L6 failures to a greater extent.

\paragraph{L7: Agent Interpretation Gap} The ChIP Atlas tool has a software, MACS2, to process raw sequencing data to identify regions which are functionally enriched. The statistical significance is computed using a logarithmic formula (-10 * log10(Q-value)), as established in the tool’s documentation. When the agent was provided with a valid BED data via the wrapper, it misinterpreted the score to be a generic confidence score, failing to apply the logarithmic definition. KEGG’s module states that a space or a plus sign, representing a connection in the pathway or the molecular complex, is treated as an AND operator and a comma, used for alternatives, is treated as an OR operator. And the module completeness requires that every top-level parenthetical block (AND-connected step) have at least one satisfied member (OR-connected alternative) present. The agent correctly identifies the boolean 'OR' behavior of commas in the module definition but misinterprets the module completeness rule by stating that missing a single K-number within a group does not compromise functionality, ignoring the fact that module completeness requires at least one present K-number for every parenthetical group (step) in the module definition. Representative instances linking reliability dimensions to failure loci are provided in Appendix~\ref{app:dimension-locus-examples} (Table~\ref{tab:dimension-locus-examples}).

\section{Discussion}
\paragraph{Why silent failures occur?} This study was designed to identify and characterize silent failures in agent--tool interactions, rather than to establish their engineering or organizational root causes. We therefore treat the following as plausible ecosystem-level contributors, not as causal explanations for every observed case. First, rapidly evolving agent architectures, models, APIs, and tool integrations can outpace systematic testing of cross-component behaviors; recent work on multi-agent systems similarly identifies specification and system-design failures as important sources of failure \citep{cemri2025mast}. Second, information flow is distributed across tool providers, API maintainers, wrapper developers, and agent developers, creating coordination and dependency-stability challenges \citep{bogart2015dependencies,lercher2024apievolution}. Third, scientific resources and their APIs evolve through updates, revisions, and changing documentation; empirical studies of Web and microservice APIs show that versioning, change-impact analysis, and communicating changes to consumers remain persistent challenges \citep{sohan2015webapi,lercher2024apievolution}. Finally, when an API or community-maintained wrapper exposes only part of a tool's Web functionality, the integration can remain technically functional while still being incomplete for an agent's intended task. These factors motivate evaluating reliability across the complete Agent--Wrapper--API--Tool chain rather than treating any single component as the sole source of failure.

\paragraph{What does it mean?} Silent failure is a property of an interaction pathway: the relevant unit of analysis is not an API, wrapper, or agent in isolation, but the complete Agent--Wrapper--API--Tool chain. Provenance research similarly treats the origin, transformation, and movement of information across a workflow as essential to interpreting its outputs \citep{buneman2001why,simmhan2005survey,davidson2008provenance,johns2023provenance}. A silent failure therefore cannot always be attributed to the underlying tool alone. Nor does a successful tool call establish that the agent has completed the task correctly or used the returned result faithfully; diagnostic agent evaluations distinguish, for example, calling a needed tool from ignoring its result or fabricating an unsupported output \citep{qin2024toolllm,liu2024agentbench,soni2026toolfailbench}.

Consequently, an audit must establish whether all information required for the task was retrieved, whether relevant qualifiers and metadata were retained, and whether the output supports the intended scientific purpose. This is particularly important in biological workflows, where provenance, context, and metadata shape the appropriate reuse and interpretation of data \citep{wilkinson2016fair,johns2023provenance}. Unlike explicit errors, partial results remain available for downstream reasoning. Omitted qualifiers---such as evidence type, organism specificity, or search scope---can therefore alter an agent's interpretation without preventing execution. Truncation, unavailable filtering, missing metadata, and ambiguous semantics can each propagate through the interaction chain and be amplified into a stronger downstream claim than the retrieved evidence supports.

\paragraph{Why does this matter?} Silent failures have four important implications. First, scientific workflows often compose multiple tools, so a limitation at one stage can propagate into later retrieval, analysis, or interpretation unless its provenance is retained \citep{simmhan2005survey,davidson2008provenance,cuevas2013workflows}. Second, verifying an apparently successful agent output can impose a substantial burden on users, particularly when the system appears reliable enough to encourage over-reliance on automated recommendations \citep{parasuraman1997humans,goddard2012automation}. Third, an apparent efficiency benefit can be offset by downstream checking, reanalysis, or rework when a limitation is discovered only after its output has been reused; provenance and workflow-reproducibility research emphasizes the importance of tracing such divergences to support diagnosis and reuse \citep{missier2014provenance,davidson2008provenance}. Fourth, a silent failure may yield a fluent, coherent answer without establishing factual correctness, faithfulness to retrieved information, or adequacy of evidence for the stated conclusion \citep{ji2023hallucination}. 

\paragraph{What can be done?} We believe that users cannot reasonably be expected to understand the implementation and limitations of every tool available in an expanding agentic ecosystem. While we believe skepticism is needed for the users while they leverage agentic pipelines for scientific workflows, we foresee seven clear directions as future work in this space. (1) Develop measurable contextual reliability indicators rather than binary tool-success measures. (2) Develop automated tests for qualifier preservation; completeness; provenance; scope; semantic preservation; and limitation disclosure. (3) Develop runtime triggers and disclosures that alert agents/users when an interaction crosses a known reliability boundary. Tool limitations should become part of the agent's evidence state. (4) Develop mechanisms for propagating tool limitations through wrappers and agents so that they remain visible in the final scientific output. (5) Enable testing and evaluation of tool configurations at the time of deployment and periodically thereafter. The testing shall cover API availability, parameter mapping, filed preservation, filtering, pagination, output formats, error handling, semantic interpretation etc. (6) Encourage tools and wrappers to explicitly disclose (in machine-readable formats) the functionalities that are unavailable, differences between web and API interfaces, logics and ranking mechanisms adopted in the API, unsupported parameters, etc. (7) Establish tool ecosystem needs monitoring mechanism. A bug, failure or even adversary in one of the tools can impact the other tools in a given task or a set of tasks. Also, given the tools are managed decentrally having a collective view on periodic maintenance, bug resolutions and updates can become relevant indicators for reliability. 

\emph{Our proposition of contextual reliability:}  Contextual reliability is the extent to which an Agent--Tool interaction preserves and communicates the information, qualifiers, provenance, scope, and meaning required to support the intended scientific conclusion, across the interaction chain. Unlike conventional tool reliability, which may focus on whether a tool call executes successfully or returns valid data, contextual reliability evaluates whether the returned information remains sufficient and correctly interpretable in context (e.g a response has low contextual reliability if relevant information is truncated, qualifiers are lost, provenance is unavailable, scope is unclear, or the agent assigns a meaning not supported by the retrieved evidence). 

\section{Limitations and conclusion}

We recognize that our study is not exhaustive and has certain limitations. This study evaluates a ToolUniverse-specific environment, and deep analysis covers 15 main tools (having 104 associated ToolUniverse wrappers) rather than the entire tool ecosystem. Our approach to candidate generation and primary adjudication use LLMs, while manual validation covers only a subset of cases. While two authors have reviewed the outcomes of the audit results and reperformed tests, the candidate generation may have missed some aspects which leverged LLM for such task. The APIs, wrappers, interfaces, and data evolve, so individual observations may change with tool versions, its not static and our tests may vary with the changes. Establishing that a failure is silent is inherently harder than detecting an explicit error, we have approached it in a certain way and our approach is not exhaustive. The results are an empirical characterization of observed risks, not an exhaustive estimate of all agent--tool failures and as stated earlier in our approach the candidate generation may influence the proportion of failures observed and hence proportion of failures are not to be considered as representative of typical silent failures in agent-tool interaction.

In this study, we empirically characterize silent failures in agent–tool interactions across 15 scientific tools and 198 candidate test cases. We introduce a failure-locus taxonomy spanning tool limitations, API gaps, wrapper gaps, agent usability, and agent interpretation. We identified recurring mechanisms through which partial or transformed tool outputs become downstream scientific claims, including quiet subset presentation and silent amplification and we propose contextual reliability as a mechanism for evaluating whether information, qualifiers, provenance, scope, and meaning remain sufficient for the intended scientific conclusion. We contribute these via a focused study of tools integrated within ToolUniverse. We would like to clarify that ToolUniverse serves as our experimental environment rather than the object of the study itself. We use its collection of scientific tools to systematically examine failure modes that can arise in agent–tool interactions.

\bibliographystyle{plainnat}
\begingroup
\raggedright
\bibliography{references}
\endgroup

\appendix
\section{Failure Measurement}
\label{app:rubrics}

\begingroup
\scriptsize
\setlength{\tabcolsep}{3pt}
\renewcommand{\arraystretch}{1.08}
\begin{longtable}{>{\raggedright\arraybackslash}p{0.17\textwidth}>{\raggedright\arraybackslash}p{0.27\textwidth}>{\raggedright\arraybackslash}p{\dimexpr0.56\textwidth-6\tabcolsep\relax}}
\caption{Dimensional classification of failures and rubrics.}\label{tab:rubrics}\\
\toprule
Measure & Dimensionality & Key checks / verification requirements \\
\midrule
\endfirsthead
\toprule
Measure & Dimensionality & Key checks / verification requirements \\
\midrule
\endhead
\bottomrule
\endfoot
1. Completeness & Information, field, record, and qualifier coverage & Check whether all requested information is present; whether fields, records, references, qualifiers, alternatives, or pagination have been omitted; whether a response that claims completeness is actually complete; compare API/tool output $\rightarrow$ wrapper output $\rightarrow$ agent response. \\
\midrule
2. Correctness & Factual, identifier, logical, and semantic accuracy & Verify values against authoritative sources; check identifiers, mappings, calculations, and claims; check whether the agent confused entities or interpreted a tool field incorrectly; distinguish retrieved facts from inferred facts. \\
\midrule
3. Appropriateness & Tool, endpoint, and parameter selection; query construction; scope fit & Check whether the correct tool/endpoint was selected; whether the tool can answer the question; whether parameters and syntax are appropriate; whether a narrow tool was incorrectly used as a comprehensive source; whether fallback tools change the intended scope. \\
\midrule
4. Consistency & Repeatability; cross-query, cross-source, and temporal consistency & Run equivalent queries repeatedly; compare alternative formulations and reverse-direction queries; test ordering, batching, and namespace transformations; determine whether differences are explainable by source semantics or execution conditions. \\
\midrule
5. Reliability & Execution stability, error handling, recovery, and dependency behavior & Check for timeouts, errors, empty results, truncation, and partial responses; determine whether failures are surfaced; test whether retries/fallbacks preserve scope; check whether apparent success represents failed or partial execution. \\
\midrule
6. Attribution & Source and evidence provenance, citation completeness, and traceability & Check whether claims can be traced to sources; whether identifiers, references, evidence codes, and source records survive; whether the tool $\rightarrow$ wrapper $\rightarrow$ agent chain can be reconstructed; whether citations support the specific claims. \\
\midrule
7. Interpretation & Source neutrality, ranking effects, hidden assumptions, and confidence/calibration & Check whether ranking or selection logic is exposed; whether defaults or hidden weighting affect results; whether ranked results become factual conclusions; whether uncertainty, confidence, or evidence strength is preserved; check for subjective framing. \\
\midrule
8. Relevance & Semantic meaning, qualifier preservation, contextual interpretation, and inference & Check whether fields and symbols are correctly interpreted; whether qualifiers such as evidence, status, species, confidence, direction, temporal state, or experimental/computational origin survive; whether qualified claims become absolute; distinguish source content from agent inference. \\
\midrule
9. Context \& conditions & Query alignment, information relevance, scope relevance, and downstream relevance & Check whether the response answers the question asked; whether relevant records are included; whether irrelevant information replaces requested information; whether the answer remains at the requested specificity; whether narrowing or substitution is disclosed. \\
\midrule
10. Purpose alignment & Intended-use alignment, contextual fitness, downstream impact, and purpose evolution & Check whether the output suits the intended use; whether tool limitations affect that use; whether a technically valid answer could produce a misleading downstream conclusion; check changes in tool/API purpose or scope where relevant. \\
\end{longtable}
\endgroup

\section{Dimensions, Failure Loci, and Representative Instances}
\label{app:dimension-locus-examples}

\begingroup
\small
\setlength{\tabcolsep}{4pt}
\renewcommand{\arraystretch}{1.1}
\begin{longtable}{@{}>{\raggedright\arraybackslash}p{0.16\linewidth}>{\raggedright\arraybackslash}p{0.21\linewidth}>{\raggedright\arraybackslash}p{\dimexpr0.63\linewidth-4\tabcolsep\relax}@{}}
\caption{Dimensions, associated failure loci, and representative instances.}\label{tab:dimension-locus-examples}\\
\toprule
\textbf{Dimension} & \textbf{Associated Failure Loci (indicative)} & \textbf{Instances} \\
\midrule
\endfirsthead
\toprule
\textbf{Dimension} & \textbf{Associated Failure Loci (not exhaustive)} & \textbf{Instances} \\
\midrule
\endhead
\endfoot
\bottomrule
\endlastfoot
Completeness & L7, L5, L3 & 1. ClinicalTrials.gov (L7, CTG\_TC\_02): Agent misformulated an input query leading to missing information on FDAAA 801 violations.
  \par\smallskip 2. AMPSphere (L5, AMPS\_TC\_01): Absence of co-prediction scores available in the API baseline. 
  \par\smallskip 3. BIGG Models (L3, BIGG\_TC\_7): The BiGG web search interface provides a functional 'multistrain=off' filter, but this capability is missing from the API endpoints and documentation.
  \\
\midrule
Interpretation Variabilities & L7, L5 & 1. KEGG (L7, KEGG\_TC\_15): Agent correctly parses OR-logic within a module group but misapplies the AND-completeness rule across blocks.
  \par\smallskip 2. BV-BRC (L5, BVBRC\_TC\_4): API returns the critical quality metrics ('checkm\_completeness', 'checkm\_contamination', 'genome\_quality'), as confirmed by the tool\_api\_response. 'BVBRC\_get\_genome' wrapper fails to include these fields in its return schema -- omits them from the actual data returned to the agent.
  \par\smallskip 3. KEGG (L7, KEGG\_TC\_15): The agent correctly identifies the boolean 'OR' behavior of commas in the module definition but misinterprets the module completeness rule by stating that missing a single K-number within a group does not compromise functionality, ignoring the fact that module completeness requires at least one present K-number for every parenthetical group (step) in the module definition. This is a misinterpretation by the agent of the logic provided by the wrapper/API.
  \\
\midrule
Result Appropriateness & L7, L5, L3 & 1. ChIP-Atlas (L7, CHIPATLAS\_TC\_08): Agent misreads MACS2's -10\ensuremath{\times}log10(Q-value) score as a raw/linear magnitude.
  \par\smallskip 2. AMPSphere (L5, AMPS\_TC\_04): Undocumented whether exact-match search surfaces homology-based hits from the separate MMseqs/HMMER endpoints.
  \par\smallskip 3. Expression Atlas (L3, GXA\_TC\_10): Missing data description critical for inferring values: The gene expression value is returned without mentioning the unit (TPM/FKPM).
   \\
\midrule
Source and Data Attribution & L3, L1, L5  & 1. BiGG Models (L3, BIGG\_TC\_04): PMID reference fields exist in schema but are empty for specific named models.
  \par\smallskip 2. AMPSphere (L1, AMPS\_TC\_08): "Progenomes" --- a genome database source, not an environmental location --- is incorrectly classified as a habitat value alongside genuine habitats like "human gut."
  \par\smallskip 3. BV-BRC (L5, BVBRC\_TC\_03): Data fields like assembly\_accession, bioproject\_accession and p2\_genome\_id are missing from the Wrapper.
  \\
\midrule
Ranking and Relevance & L5, L3 & 1. AMPSphere (L5, AMPS\_TC\_04): MMseqs2/HMMER endpoints are present in the API baseline but missing from the wrapper. 
  \par\smallskip 2. ClinicalTrials.gov (L3, CTG\_TC\_10): Web UI defaults to "Relevance" sort; API default sort order differs/is unclear.
  \par\smallskip 3. EMDB (L5, EMDB\_TC\_09): The API implementation enforces an undocumented or inconsistent result cap (e.g., 100 or 1000) that does not match the needs of a repository with 61,000+ entries, preventing effective data harvesting.
   \\
\midrule
Reliability of Information & L5, L3 & 1. Expression Atlas (L3, GXA\_TC\_05): Accession-naming drift between E-MTAB and E-GTEX conventions over time creates evolutionary reliability risk for long-running studies.
  \par\smallskip 2. BV-BRC (L5, BVBRC\_TC\_12): Legacy PATRIC-era p2\_genome\_id retained for cross-walks may drift from current BV-BRC taxonomy. 
  \par\smallskip 3. InterPro (L3, IPR\_TC\_09): The API (and thus the wrapper) lacks transparency regarding the underlying InterProScan version (5 or 6) which affects the fundamental reproducibility and meaning of sequence match data.
  \\
\end{longtable}
\endgroup

\section{Tool-wise representative examples}

\subsection{AMPSphere}

AMPSphere provides antimicrobial-peptide sequence and functional information \citep{santosjunior2024discovery}, but failures arise from ambiguous biological semantics, incomplete contextual information, and difficulties in using or interpreting returned results. The key failures observed are: \textbf{L1:} The `Blank' and `others' microbial-source categories do not provide sufficiently specific semantic definitions to support environmental interpretation. \textbf{L3:} Structure and family-related raw data available  in the web UI is missing in the API, causing the agent to miss crucial factors that might contribute to interpretaion of the information. \textbf{L7:} The agent does not consider all the quality scores before interpreting a sample as low/high quality. 

\subsection{BV-BRC}

BV-BRC provides microbial genome information whose interpretation depends on genome quality, provenance, identifier semantics, and reference status \citep{olson2023bvbrc}. The key failures observed are: \textbf{L5:} Fields like checkm scores and genome quality status is missing from the wrapper, with no way for the agent to determine agent quality. \textbf{L3:} The API baseline does not contain genome assembly information present in the web UI, causing the agent to assume a genome assembly. \textbf{L5:} The reference/representative genome was missing from the wrapper, the agent might select one as the reference without source evidence establishing that designation.

\subsection{ClinGen}

ClinGen provides curated genomic validity information \citep{clingen2025resource}, but mismatches in API functionality, output representation, and biological granularity can lead agents to answer the wrong scientific question. The key failures observed are: \textbf{L3:} Drug-name/RXNORM searching available through the Web interface is not represented as a corresponding API capability; for example, a drug-oriented search such as ``aspirin'' cannot be reproduced through the documented validity endpoints. \textbf{L5:} The gene validity search wrapper does not support genomic coordinate strings as input limiting functionality. 

\subsection{CIViC}

CIViC integrates clinical variant evidence and assertions, making evidence granularity, search coverage, and evidence-count interpretation critical for reliable agent use \citep{griffith2017civic}. The key failures observed are: \textbf{L5:} The API can flatten distinctions between individual Evidence items and synthesized Assertions, reducing the granularity of the underlying clinical evidence. \textbf{L5:} Advanced Web search supports Boolean logic and ontology expansion that are not equivalently exposed through the API; for example, a parent disease search may not automatically retrieve evidence associated with specific subtypes. 

\subsection{ChIP-Atlas}

ChIP-Atlas provides large-scale epigenomic information, but biological-state normalization, score semantics, ranking, and result retrieval introduce multiple failure loci \citep{zou2024chipatlas}. The key failures observed are: \textbf{L1:} Biologically meaningful modifications may be intentionally collapsed; for example, phospho-SMAD3 is normalized to SMAD3, removing information about the phosphorylation state. \textbf{L2:} Absence of API endpoint to run differential analysis. \textbf{L3:} Search and ranking semantics are incompletely specified. \textbf{L7:} The agent interpreted a MACS2 score of 500 as a raw count rather than the documented $-10\log_{10}(q)$ transformation. 

\subsection{CryoET Data Portal}

CryoET Data Portal provides cryo-electron tomography datasets \citep{ermel2024dataportal}, but provenance, attribution, and retrieval metadata may not survive consistently across the automated interface. The key failures observed are: \textbf{L3:} Publication metadata may be returned as null; for example, \texttt{datasetPublications} can be absent despite publication context associated with the dataset. \textbf{L7:} The official GraphQL API supports retrieving child datasets for a deposition (evidenced by the API response which successfully returns datasets linked to deposition 10014), but the CryoET\_list\_depositions wrapper completely omits this field from its return schema, preventing the agent from discovering the hierarchical relationship.

\subsection{EMDB}

EMDB provides structural biology records through identifiers and searchable metadata \citep{wwpdb2024emdb}, but retrieval limits and identifier representations can produce failures that propagate into agent interpretation. The key failures observed are: \textbf{L5:} Search results are capped; for example, the \texttt{rows} parameter is limited to 1000 despite the database containing substantially more records, creating systematic truncation for broad searches. \textbf{L3:} The official documentation for the tool's advanced search interface specifies dozens of filterable fields, but the actual implementation of the REST API (as verified by the failure of the fielded queries in the baseline test and the documentation's limitation to a simple keyword search endpoint) only supports basic keyword searching. This constitutes a partial functional gap where the API fails to provide the same capability as the WebUI.

\subsection{BiGG Models}

BiGG Models provides metabolic models and associated metadata \citep{king2016bigg}, but differences between Web, API, and wrapper capabilities can constrain automated model discovery. The key failures observed are: \textbf{L3:} Model metadata can differ between search and model-detail representations; for example, organism information available in the search context may be absent from the model detail response. \textbf{L3:} The Web interface supports more advanced multi-field search than the corresponding REST API. \textbf{L5:} The official Tool API successfully returns gene sequence and strand data, but the ToolUniverse wrappers do not expose these fields to the agent. While the BiGG\_get\_model wrapper exists, it only provides high-level metadata; there is no wrapper functionality to access specific gene-level details returned by the underlying API endpoint (e.g., /api/v2/models/\{model\_id\}/genes/\{gene\_id\}), forcing the agent to lose information.

\subsection{ClinicalTrials.gov}

ClinicalTrials.gov provides extensive clinical-trial search and eligibility information \citep{nlmclinicaltrials}, but API filtering limitations can change the population retrieved by an autonomous agent. The key failures observed are: \textbf{L3:} The wrapper returns a success status with empty outcomes but no explicit 'No Results Posted' flag. This is a silent failure to convey the definitive 'No Results Posted' state clearly shown in the Web UI. \textbf{L7:} Agent was asked to find studies which had "FDAAA 801 Violations". Instead of using the correct query term which is "FDAAA 801" only, it used the entire string "FDAAA 801 violations" to search, due to which the query returned empty response

\subsection{CTIS}

CTIS provides clinical-trial information \citep{emactis} through Web and API interfaces whose search semantics, status representations, and result-shaping capabilities are not fully aligned. The key failures observed are: \textbf{L7:}. Agent misinterprets completeness of results due to pagination. \textbf{L5:} The agent treated API search behavior as equivalent to the Web-interface Boolean filtering semantics, although the API did not establish equivalent behavior.

\subsection{ClinVar}

ClinVar contains submitted reports of human genomic variants for diseases and drug responses with supporting evidence \citep{landrum2014clinvar}, however, there are few API/wrapper differences that might cause misinterpretations in agents. For example, L5: Detailed submission-level observation metadata, including Age, Sex, Clinical Features, and Collection Method, is present in the underlying `efetch` XML but omitted by the ToolUniverse wrapper, L5: The Web interface distinguishes submissions that contribute to the aggregate classification from those that do not using `C`/`N` badges, but this contribution status is not exposed by the wrapper.

\subsection{Expression Atlas}

Expression Atlas is an open source resource providing gene and protein expression data for different biological conditions \citep{madrigal2026expressionatlas}. Some inconsistencies and incompleteness in information through API/wrapper might hinder agentic workflows. L3: The Expression Atlas Web UI provides gene-level expression values (TPM/FPKM), but the documented API exposes only experiment-level metadata and no equivalent programmatic endpoint for retrieving the full expression data.  L3: The Web UI exposes curated collections such as ENCODE and BLUEPRINT, but the API's `experimentProjects` field can be empty even for experiments belonging to these projects, preventing reliable programmatic identification of project affiliations. L3: Experiment accession identifiers are inconsistent across the Web interface and API (e.g., E-MTAB-5214 vs E-GTEX-8), and the ToolUniverse tool fails to resolve the former or provide alias mapping. L3: The Web UI provides publication/citation metadata, but the experiment API omits PubMed IDs and other citation fields. L3: Expression values lack explicit TPM/FPKM unit metadata in the API, requiring the agent to infer the units rather than receiving them as a qualifier. 

\subsection{InterPro}

Interpro provides functional analyses of proteins by classifying protein sequences into families and predicting the presence of important sites \citep{hunter2009interpro}, in which certain differences in features and capabilities are partially or completely missing in the API/wrapper. L3: The InterPro Web interface provides a `Search by sequence'' capability powered by InterProScan, while the official InterPro REST API does not have a similar endpoint for submitting raw protein sequences for scanning, preventing agents from performing on-the-fly domain classification through the core API. L3: The Web interface supports `Search by Domain Architecture'' to identify proteins matching ordered combinations of domains, while the API does not provide an equivalent database-wide architecture-pattern search, limiting automated discovery of proteins with specific domain arrangements. L3: The Web interface displays full taxonomic lineage on protein pages, whereas the API response exposes limited organism information and requires additional taxonomy requests to reconstruct the lineage available directly to Web users. 

\subsection{KEGG}

The KEGG database and analysis tools enable understanding high-level functionalities of biological systems including cells, organisms, ecosystems from molecular information \citep{kanehisa2000kegg}. However absence of certain functionalities restrict agentic workflows from utlizing the complete capabilities of KEGG. L4: High-level analytical operations available on the Web (KEGG Mapper's Reconstruct/Color/Join, and sequence-based functional annotation via BlastKOALA/GhostKOALA) are not exposed through the REST API, which offers only raw lookup, list, and conversion operations. L2: The API cannot reproduce the full analytical and annotation capability of the Web interface programmatically, and where richer machine-readable structure does exist (e.g. KGML), it is returned as unparsed, interactivity-stripped XML rather than an agent-consumable format. L7: The agent compensated for the absence of a computed module-completeness endpoint by manually interpreting the DEFINITION field's boolean syntax itself, correctly identifying comma-separated K-numbers as alternative (OR) options but incorrectly generalizing this to conclude that a missing K-number within a group never compromises module completeness.

\subsection{Reactome}

Reactome provides peer-reviewed pathway information \citep{croft2011reactome} through Web and API interfaces whose hierarchical relationships, identifier semantics, search behaviour, and analysis metadata are not fully aligned. The key failures observed are:
L3: The Reactome Content Service does not provide a per-instance `referrers'' endpoint or equivalent field, requiring agents to reconstruct parent relationships through child `eventOf'' links and potentially perform N+1 queries, limiting complete traversal of the pathway hierarchy. L3: Species identifiers are inconsistent across Reactome services where Content Service operations may use NCBI taxonomy IDs such as `9606'', whereas Analysis Service operations such as species comparison require Reactome-specific `dbId'' values such as `48892'', forcing agents to translate between two identifier systems for the same species. L3: Reactome search exposes clustering as a default behaviour while the corresponding ToolUniverse search interface does not expose a `cluster'' parameter, and the API enforces a 200-character limit on the `q'' parameter, restricting complex automated searches. L1: Reactome's use of `Reaction'' encompasses a broad range of biological state changes, including binding and translocation events, which may lead agents to interpret Reactome reaction records as conventional biochemical conversions when they are not. 

\end{document}